\documentclass{article}

\usepackage[preprint]{neurips_2025}

\usepackage[utf8]{inputenc} 
\usepackage[T1]{fontenc}    
\usepackage{hyperref}       
\usepackage{url}            
\usepackage{booktabs}       
\usepackage{amsfonts}       
\usepackage{amsmath}
\usepackage{nicefrac}       
\usepackage{microtype}      
\usepackage{xcolor}         
\usepackage{graphicx}
\usepackage{subcaption}
\usepackage{wrapfig}
\usepackage{sidecap}
\usepackage{soul}

\title{Learned Image Compression for Earth Observation:\\Implications for Downstream Segmentation Tasks}
\workshoptitle{Learned Image Compression for Earth Observation:\\Implications for Downstream Segmentation Tasks}

\author{%
  Christian Mollière\\
  OroraTech GmbH\\
  Munich\\
  \And
  Iker Cumplido\\
  Department of Informatics\\
  LMU, Munich\\
  \And
  Marco Zeulner\\
  Department of Informatics\\
  LMU, Munich\\
  \And
  Lukas Liesenhoff\\
  OroraTech GmbH\\
  Munich\\
  \And
  Matthias Schubert\\
  Department of Informatics\\
  LMU, Munich\\
  \And
  Julia Gottfriedsen\\
  OroraTech GmbH\\
  Munich\\
}

\begin{document}

\maketitle

\begin{abstract}
The rapid growth of data from satellite-based Earth observation (EO) systems poses significant challenges in data transmission and storage. We evaluate the potential of task-specific learned compression algorithms in this context to reduce data volumes while retaining crucial information. In detail, we compare traditional compression (JPEG 2000) versus a learned compression approach (Discretized Mixed Gaussian Likelihood) on three EO segmentation tasks: Fire, cloud, and building detection. Learned compression notably outperforms JPEG 2000 for large-scale, multi-channel optical imagery in both reconstruction quality (PSNR) and segmentation accuracy. However, traditional codecs remain competitive on smaller, single-channel thermal infrared datasets due to limited data and architectural constraints. Additionally, joint end-to-end optimization of compression and segmentation models does not improve performance over standalone optimization. 
\end{abstract}

\section{Introduction}
Satellite-based Earth Observation (EO) systems have become essential for monitoring and managing environmental phenomena, particularly wildfires, which pose severe threats to ecosystems and human communities. However, the expanding spatial coverage and increased resolution from EO missions create significant challenges in terms of data transmission and storage. To address these challenges, well-designed image compression algorithms are key. Unlike traditional compression methods, which primarily seek to preserve visual fidelity, task-specific compression aims to retain essential features necessary for accurate downstream analysis, such as image segmentation. Effective task-specific compression would allow significant data volume reductions while maintaining the performance and interpretability of critical tasks like wildfire and cloud detection.  The research question that motivates our work is therefore: What is the optimal EO compression regime for various downstream tasks, while maintaining a balanced rate-distortion trade-off? In this paper, we compare the performance of traditional image compression techniques, specifically JPEG 2000, with that of learned compression approaches across three representative EO segmentation tasks: cloud detection, fire detection, and building footprint extraction. By systematically evaluating these methods on diverse datasets differing in spectral modality (thermal infrared versus optical multispectral), dataset size, and spatial resolution, we identify optimal compression strategies tailored to specific EO scenarios.

\begin{figure}[ht]
  \centering
    \begin{subfigure}[b]{0.4\textwidth}
    \includegraphics[width=\linewidth]{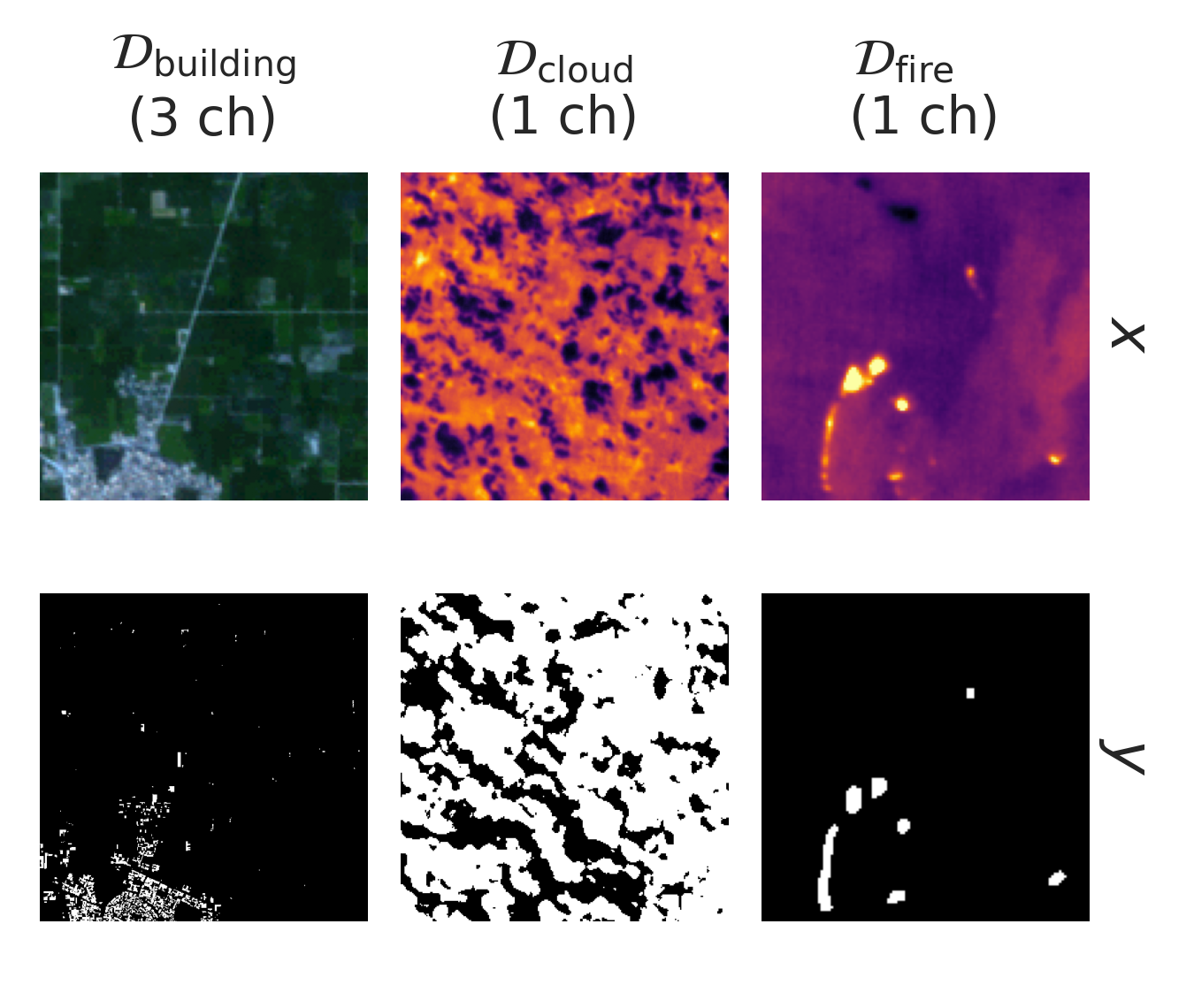}
    \caption{Samples of the datasets used in this work. The thermal segmentation tasks are evaluated monochromatically, whereas the optical task is providing three channels.}
    \label{fig:dataset}
  \end{subfigure}
  \hfill
  \begin{subfigure}[b]{0.55\textwidth}
    \includegraphics[width=\linewidth]{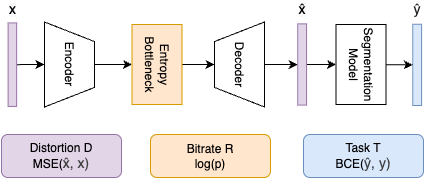}
    \vspace{1.1em}
    \caption{Overview of experiment setup for task-specific raster data compression. Raw observations $x$ are compressed onboard the satellite and decompressed at the ground station for segmentation analysis.}
    \label{fig:architecture}
  \end{subfigure}
\end{figure}

\vspace{-1em}

\section{Experimental Setup}
\label{sec:exp_setup}
\paragraph{Datasets}
We conducted experiments using three diverse EO datasets representing distinct segmentation tasks: Cloud detection, fire detection, and building footprint extraction. Collectively, these datasets allow comprehensive evaluation of compression strategies across various EO scenarios, ensuring our findings are suitable for real-world applications. Samples are shown in \subref{fig:dataset}.

\begin{itemize}
    \item \textbf{Fire Dataset} (\cite{roetzer2025selfsupervised})
The fire dataset $\mathcal{D}_{\text{fire}}$ originates from a commercial CubeSat platform using mid-wave infrared (MWIR) imagery at 3.8 µm wavelength, optimal for detecting active combustion. Initially comprising 40 full scenes, we augment the data by cropping 571 smaller 32×32 pixel patches around confirmed fire locations, each accompanied by expert-annotated binary fire masks.

    \item \textbf{Cloud Dataset} (\cite{wolki2024exploring})
The cloud dataset $\mathcal{D}_{\text{cloud}}$ comprises thermal infrared imagery also acquired from commercial CubeSat constellation. Each image has a ground sampling distance (GSD) of 200 m, capturing radiance in the long-wave infrared (LWIR) spectrum, centered at 11.5 µm. The dataset consists of 528 manually annotated scenes of cloud presence. Each scene is resized to 256×256 pixels.

    \item \textbf{Building Dataset} (\cite{prexl_potential_2023})
The building dataset $\mathcal{D}_{\text{building}}$ is based on multispectral optical imagery from ESA's Copernicus Sentinel-2 mission, featuring spatial resolutions of 10–20 m per pixel. We utilized 28,828 10-channel image-mask pairs, each resized to 128×128 pixels. The masks represent building footprints extracted from OpenStreetMap and Microsoft global building footprint layers, providing a substantial dataset suitable for training and evaluating the segmentation and compression models.

\end{itemize}


\paragraph{Segmentation Baseline Methods}
To establish robust baseline performance, we employ a U-Net architecture (\cite{ronneberger2015u}), a widely adopted convolutional neural network designed specifically for segmentation tasks. We use two different encoder backbones, MobileNetV2 (\cite{sandler2018mobilenetv2}) and ResNet34 (\cite{he2016deep}), selected for their proven effectiveness and computational efficiency. Models are trained with binary cross-entropy (BCE) loss and optimized using the Adam optimizer. Training incorporates data-specific augmentation and early stopping criteria based on validation performance to ensure stability and generalization.

\paragraph{Compression Algorithms}
We compare two distinct image compression methods and systematically analyze the rate-distortion trade-off of each method using standard metrics. This includes the Peak Signal-to-Noise Ratio (PSNR) and bits per pixel (bpp), across different datasets.

\begin{itemize}
    \item \textbf{JPEG 2000:} A classical, wavelet-based compression method widely used for its efficiency, scalability, and support for both lossless and lossy compression. JPEG 2000 (\cite{jpeg2000}) serves as our baseline to benchmark learned compression performance.
    \item \textbf{Discretized Gaussian Mixture Likelihood:} A variational autoencoder-based compression model employing advanced entropy modeling techniques, including Gaussian Mixture Likelihoods and attention mechanisms (\cite{cheng2020learnedimagecompressiondiscretized}). It is specifically designed to reduce spatial redundancy in its latent representation, which is of importance for efficient coding when fewer spectral channels are available. It significantly outperforms JPEG 2000 when evaluated on natural RGB images. We selected this architecture from the CompressAI framework (\cite{begaint2020compressai}), despite newer SoTA architectures being available, given its accessibility among other implementations and its proven performance over our baseline.
\end{itemize}

\paragraph{Optimization Methodology}
To investigate potential synergies between compression and segmentation, we implement a fully differentiable pipeline combining both models. This approach allows joint training, optimizing not only for reconstruction quality but also directly for segmentation performance. The combined loss function $\mathcal{L}$ consists of three terms. The common rate-distortion trade-off balances the distortion term $\mathcal{D}_{\text{MSE}}$ against the estimated bit-rate $\mathcal{R}$. For end-to-end optimization, we add the task-specific BCE segmentation loss $\mathcal{T}_{\text{BCE}}$ as a third weighted term. The model is trained using separate Adam optimizers (\cite{kingma2014adam}) for each of the three components (auto-encoder, entropy bottleneck and segmentation model), as the entropy bottleneck needs to be optimized using a lower learning rate. The task-specific weight $\gamma$ is set to zero for standalone compression experiments. An overview of the experiment architecture is shown in \subref{fig:architecture}. Further details on the used hyperparameters are given in Appendix \ref{sec:append_setup}.

\begin{equation}
\mathcal{L} = \lambda \cdot \mathcal{D}_{\text{MSE}} + \mathcal{R} + \gamma \cdot \mathcal{T}_{\text{BCE}}
\end{equation}

\vspace{-1em}

\section{Results}
\begin{figure}[ht]
  \centering

  \begin{minipage}[c]{0.4\textwidth}
    \centering
    \captionof{table}{Results of segmentation baseline models.}
    \label{tab:segmentation_baseline_results}
    \begin{tabular}{|c|c|c|c|}
    \hline
        \textbf{Task} & \textbf{Backbone} & \textbf{F1+} $\uparrow$ & \textbf{F1} $\uparrow$ \\
        \hline
         $\mathcal{D}_{\text{fire}}$ & MobileNetV2 & 0.665 & 0.830 \\
         $\mathcal{D}_{\text{cloud}}$ & ResNet34 & 0.792 & 0.857 \\
         $\mathcal{D}_{\text{building}}$ & ResNet34 & 0.448 & 0.720 \\
     \hline
    \end{tabular}
  \end{minipage}%
  \hfill
  \begin{minipage}[c]{0.44\textwidth}
    \centering
    \includegraphics[width=1.0\linewidth]{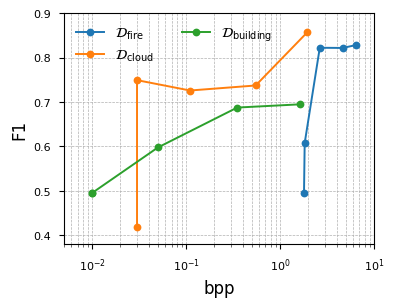}
    \captionof{figure}{Segmentation performance under bit rate reduction using JPEG 2000.}
    \label{fig:segmentation_under_bitrate_reduction}
  \end{minipage}
\end{figure}

\paragraph{Segmentation Baseline Performance}
The best performing baseline segmentation models are summarized in Table~\ref{tab:segmentation_baseline_results}. The F1+ score denotes the F1 score of the positive class. It is reported alongside its macro F1 given the class imbalance of fire and building pixels in their respective datasets. Generally, the UNet architecture is able to provide a reasonable performing baseline for all three tasks. However, it is not able to resolve very fine detail contours in the building footprints, hence the comparatively low F1+ result in $\mathcal{D}_{\text{building}}$.

\paragraph{Segmentation under Bit Rate Reduction} Figure~\ref{fig:segmentation_under_bitrate_reduction} shows the impact of bit rate reduction on segmentation performance (F1), when training on the decompressed image data. This is done to directly quantify the effect of compression on our selection of downstream tasks. For all task, we observe that a certain level of image quality is needed before reaching a plateau in segmentation performance. This is in agreement with existing work, as many segmentation tasks in EO do not seem to require the full fidelity of the used observation data \cite{GarciaSobrino2020}.

\vspace{-0.5em}


\vspace{-1em}

\paragraph{Standalone Compression Performance}

Prior to end-to-end optimization, we evaluate the standalone performance of the different lossy compression methodologies on the given set of tasks by comparing the rate-distortion tradeoff. This is commonly measured by the reconstruction (PSNR) and its bitrate per pixel (bpp).

\begin{figure}[ht]
  \centering

  \begin{subfigure}{0.45\textwidth}
    \includegraphics[width=\linewidth]{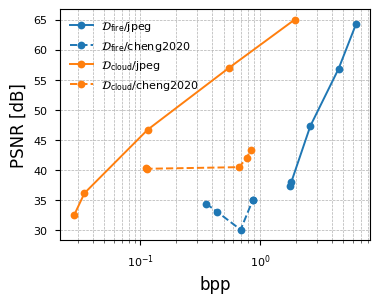}
    \caption{Results on single-channel tasks comparing JPEG2000 against the learned compression algorithm. All models were trained from scratch.}
    \label{fig:standalone_comp_fire_cloud}
  \end{subfigure}
  \hfill
  \begin{subfigure}{0.45\textwidth}
    \includegraphics[width=\linewidth]{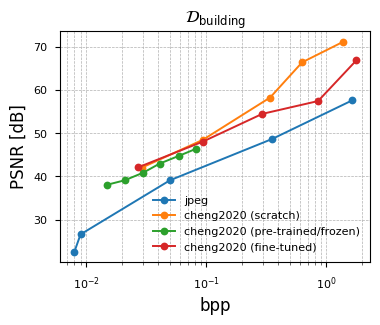}
    \caption{Results on the multi-channel task including both training from scratch and different transfer learning strategies.}
    \label{fig:standalone_comp_building}
  \end{subfigure}

\end{figure}

\begin{itemize}
    \item \textbf{Single-Channel Tasks} \subref{fig:standalone_comp_fire_cloud} shows the results on the two single-channel tasks $\mathcal{D}_{\text{fire}}$ and $\mathcal{D}_{\text{cloud}}$. JPEG2000 remains superior in the monochromatic tasks, despite testing multiple learning architectures including FullyFactorizedPrior, ScaleHyperPrior (\cite{ballé2018variationalimagecompressionscale}) and the Chenge2020Anchor (\cite{cheng2020learnedimagecompressiondiscretized}). We accredit this result due to the lack of redundant spectral information that is usually present in multi-spectral datasets. Therefore, the algorithms can only leverage the spatial context to find meaningful representations to reduce the effective entropy of the data.

    \item \textbf{Multi-channel Task} \subref{fig:standalone_comp_building} summarizes the performance on the multi-channel task $\mathcal{D}_{\text{building}}$. Here, the learned compression algorithm clearly outperforms the JPEG baseline both, when trained from scratch or fine-tuned. Additionally, the pretrained, not fine-tuned model variants from the CompressAI model zoo are shown in comparison (\cite{begaint2020compressai}). 
\end{itemize}

\vspace{-1.5em}




\vspace{-1em}

\paragraph{Task-specific Compression Performance} During our experiments, we were unable to demonstrate the benefits of joint optimization of the compression and segmentation components compared to individual optimization of the components. Details of the experiments are given in Appendix \ref{sec:end-2-end}. 


\section{Conclusion}
\label{sec:conclusion}
\paragraph{Channel-dependent Efficacy}
We observe that learned compression models substantially outperform classical codecs, such as JPEG 2000, particularly when dealing with abundant, multi-channel optical imagery datasets. The building footprint extraction task exemplifies this scenario, demonstrating clear advantages in both reconstruction quality (PSNR) and downstream segmentation accuracy. However, classical codecs remain competitive and effective for tasks involving smaller datasets or single-channel thermal imagery, such as cloud and fire detection, due to limitations in available training data and the architectural constraints of current learned compression models. This result also stood when compression and segmentation were optimized end-to-end.


\paragraph{Implications for Downstream Tasks of EO Missions}
Primarily, learned image compression methodologies work best for multi-spectral datasets with a high number of spectral features. When designing compression methodologies for datasets with low spectral count (less then three channels) we recommend starting with traditional compression codecs first. Secondly, many tasks in EO do not need the full fidelity of raw observations. Depending on the intended set of tasks lossy compression is a viable option to drastically reduce downlink overhead, while maintaining acceptable performance. Still there is a potential decrease in segmentation quality which must be weighed against the lower transfer volumes.

\paragraph{Future Directions}
Future work will aim to extend learned compression architectures specifically for single-channel EO imagery. Current methods exhibit significant limitations when applied to datasets such as thermal infrared imagery, which hinders their practical deployment. To address these challenges, it will be essential to collect larger, well-annotated datasets and to explore compression techniques that more effectively capture the spatial correlations inherent in the data.


\section{Impact Statement}
\label{sec:impact}
This paper addresses learned compression algorithms and their impact on downstream applications. This is of interest to any edge-based imaging system in EO but also other domains. Probing new emerging algorithmic solutions have to potential to significantly ease access to larger remote sensing constellations in space as they can operate under stricter mass, power and budget constraints. However, many of these technologies are of dual use. Therefore, potential advances might also be used by the defense \& intelligence community, which lie outside of the author's governance.

\bibliographystyle{plainnat}
\bibliography{bib}

@inproceedings{wolki2024exploring,
  title={Exploring Machine Learning for Cloud Segmentation in Thermal Satellite Images of the FOREST-2 Mission},
  author={W{\"o}lki, Niklas and Kondmann, Lukas and Molli{\`e}re, Christian and Langer, Martin and Werner, Martin and Gottfriedsen, Julia},
  booktitle={Proceedings of SPAICE2024: The First Joint European Space Agency/IAA Conference on AI in and for Space},
  pages={362--366},
  year={2024}
}

@inproceedings{ronneberger2015u,
  title={U-net: Convolutional networks for biomedical image segmentation},
  author={Ronneberger, Olaf and Fischer, Philipp and Brox, Thomas},
  booktitle={Medical image computing and computer-assisted intervention--MICCAI 2015: 18th international conference, Munich, Germany, October 5-9, 2015, proceedings, part III 18},
  pages={234--241},
  year={2015},
  organization={Springer}
}

@article{kingma2014adam,
  title={Adam: A method for stochastic optimization},
  author={Kingma, Diederik P and Ba, Jimmy},
  journal={arXiv preprint arXiv:1412.6980},
  year={2014}
}

@inproceedings{prexl_potential_2023,
	title = {The Potential of Sentinel-2 Data for Global Building Footprint Mapping with High Temporal Resolution},
	url = {https://ieeexplore.ieee.org/document/10144166},
	doi = {10.1109/JURSE57346.2023.10144166},
	eventtitle = {2023 Joint Urban Remote Sensing Event ({JURSE})},
	pages = {1--4},
	booktitle = {2023 Joint Urban Remote Sensing Event ({JURSE})},
	author = {Prexl, Jonathan and Schmitt, Michael},
	urldate = {2025-06-14},
	date = {2023-05},
        year = {2023},
	note = {{ISSN}: 2642-9535},
}

@ARTICLE{jpeg2000,
  author={Taubman, D.S. and Marcellin, M.W.},
  journal={Proceedings of the IEEE}, 
  title={JPEG2000: standard for interactive imaging}, 
  year={2002},
  volume={90},
  number={8},
  pages={1336-1357},
  doi={10.1109/JPROC.2002.800725}}

@article{begaint2020compressai,
	title={CompressAI: a PyTorch library and evaluation platform for end-to-end compression research},
	author={B{\'e}gaint, Jean and Racap{\'e}, Fabien and Feltman, Simon and Pushparaja, Akshay},
	year={2020},
	journal={arXiv preprint arXiv:2011.03029},
}

@misc{cheng2020learnedimagecompressiondiscretized,
      title={Learned Image Compression with Discretized Gaussian Mixture Likelihoods and Attention Modules}, 
      author={Zhengxue Cheng and Heming Sun and Masaru Takeuchi and Jiro Katto},
      year={2020},
      eprint={2001.01568},
      archivePrefix={arXiv},
      primaryClass={eess.IV},
      url={https://arxiv.org/abs/2001.01568}, 
}

@misc{ballé2018variationalimagecompressionscale,
      title={Variational image compression with a scale hyperprior}, 
      author={Johannes Ballé and David Minnen and Saurabh Singh and Sung Jin Hwang and Nick Johnston},
      year={2018},
      eprint={1802.01436},
      archivePrefix={arXiv},
      primaryClass={eess.IV},
      url={https://arxiv.org/abs/1802.01436}, 
}

@inproceedings{sandler2018mobilenetv2,
  title={Mobilenetv2: Inverted residuals and linear bottlenecks},
  author={Sandler, Mark and Howard, Andrew and Zhu, Menglong and Zhmoginov, Andrey and Chen, Liang-Chieh},
  booktitle={Proceedings of the IEEE conference on computer vision and pattern recognition},
  pages={4510--4520},
  year={2018}
}

@inproceedings{he2016deep,
  title={Deep residual learning for image recognition},
  author={He, Kaiming and Zhang, Xiangyu and Ren, Shaoqing and Sun, Jian},
  booktitle={Proceedings of the IEEE conference on computer vision and pattern recognition},
  pages={770--778},
  year={2016}
}

@article{GarciaSobrino2020,
  title = {Competitive Segmentation Performance on Near-Lossless and Lossy Compressed Remote Sensing Images},
  volume = {17},
  ISSN = {1558-0571},
  url = {http://dx.doi.org/10.1109/LGRS.2019.2934997},
  DOI = {10.1109/lgrs.2019.2934997},
  number = {5},
  journal = {IEEE Geoscience and Remote Sensing Letters},
  publisher = {Institute of Electrical and Electronics Engineers (IEEE)},
  author = {Garcia-Sobrino,  Joaquin and Pinho,  Armando J. and Serra-Sagrista,  Joan},
  year = {2020},
  month = may,
  pages = {834–838}
}

@inproceedings{roetzer2025selfsupervised,
  author = {Matthias R{\"o}tzer and Lukas Liesenhoff and Max Bereczky and Martin Ickerott and Jayendra Chorapalli and Julia Gottfriedsen},
  title = {Self-Supervised Learning for Fire Segmentation in Forest-2 images},
  booktitle = {ESA-NASA International Workshop on AI Foundation Model for EO},
  address = {Frascati, Italy},
  organization = {ESA-ESRIN},
  year = {2025},
  month = {May},
  pages = {},
}

\newpage
\appendix
\section{Hyperparameter Setup of Experiments}
\label{sec:append_setup}

\paragraph{Input Normalization} For each experiment we normalized the dataset by dividing by the global maximum of the input data to yield a distribution between [0,1].

\paragraph{Hyperparameter Tuning.}
We conducted a systematic hyperparameter search to find the optimal configuration for the Mixed Gaussian Likelihood model. The search space explored the following key hyperparameters using grid search.

\begin{itemize}
    \item \textbf{Quality Level ($q$):} The \texttt{CompressAI} library offers discrete quality levels from 1 to 6. These levels primarily control the network's capacity by setting the dimensionality of the latent space. We observed that levels $q \in \{1, 2, 3\}$ corresponded to a latent dimension of 128, while levels $q \in \{4, 5, 6\}$ used a dimension of 192. To represent these two main configurations, we focused our final experiments on levels 3 and 6.
    \item \textbf{Batch Size:} We explored batch sizes in the set $\{4, 8, 16, 32\}$.
    \item \textbf{Learning Rates ($\eta$):} Separate learning rates were tuned for the main network ($\eta$) and the auxiliary hyperprior network ($\eta_{\mathrm{aux}}$), both within the range of $[10^{-6}, 10^{-1}]$.
    \item \textbf{Rate-Distortion Weight ($\lambda$):} This crucial hyperparameter balances the trade-off between the rate and distortion terms in the loss function. We explored values in the range $[10^{-4}, 10^{3}]$.
\end{itemize}

\paragraph{Loss Function and Optimizer.}
The training objective for the Mixed Gaussian Likelihood model is the rate-distortion loss function described in Section \ref{sec:exp_setup}. This function combines a rate term, encouraging compact representations, with a distortion term, enforcing reconstruction accuracy. We employed the Adam optimizer to train the compression and segmentation model s for all experiments (\cite{kingma2014adam}).

\paragraph{Computing Resources} All training has been done using a single NVIDIA RTX 4000. Training times were usually in the range from hours to days.

\section{Results of Joint Optimization}
\label{sec:end-2-end}
We focus on $\mathcal{D}_{\text{building}}$ for the evaluation of end to end optimization, given the apparent unsuitability of monochromatic tasks for learned compression algorithms. The results prior and post optimization are summarized in Table~\ref{tab:end2end}. In summary, end-to-end optimization is able to converge both the compression and the segmentation components when starting from low quality models. However, the final performance does not exceed standalone optimization of the individual components. 

The used loss weights for these results are ($\lambda=10.0$, $\gamma=1e-3$) for the high-quality start and ($\lambda=10.0$, $\gamma=1e-2$) for the low-quality start respectively. The start scenarios are initialized using pretrained components for both, compression and segmentation. In the high-quality start scenario, the compression model has been optimized for quality, whereas the low-quality scenario uses a very lossy compression.

\begin{table}[h]
    \centering
    \caption{Performance of experiments prior and post to end-to-end optimization.}
    \vspace{0.5em}
    \begin{tabular}{|c|c|c|c|c|c|c|}
    \hline
        \textbf{Scenario} & \textbf{bpp$_{\text{prior}}$} & \textbf{PSNR$_{\text{prior}}$} [dB] $\uparrow$ & \textbf{F1$_{\text{prior}}$} $\uparrow$ & \textbf{bpp} & \textbf{PSNR} [dB] $\uparrow$ & \textbf{F1} $\uparrow$ \\
        \hline
         high-quality start & 0.6121 & 66.29 & 0.7281 & 0.6106 & 66.64 & 0.7238 \\
         low-quality start & 8e-5 & 25.76 & 0.4925 & 0.4407 & \textbf{49.80} & \textbf{0.7090} \\
     \hline
    \end{tabular}
    \label{tab:end2end}
\end{table}

Generally, we explored a range of different hyperparameters for the joint optimisation.

\begin{itemize}
    \item \textbf{Rate-Distortion Weight ($\lambda$):} $\{0.1, 1, 10, 100\}$.
    \item \textbf{Task-specific Weight ($\gamma$):} $\{0.001, 0.01, 0.05, 0.1, 1\}$.
\end{itemize}

\end{document}